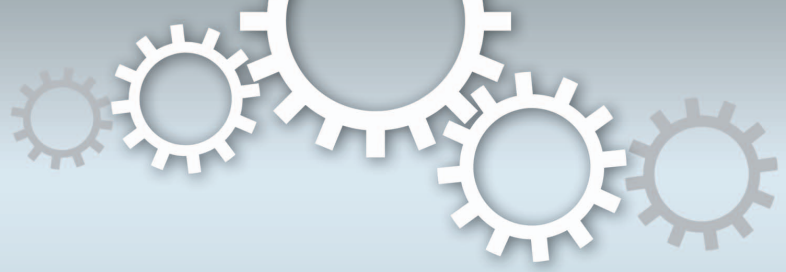

# GBM Volumetry using the 3D Slicer Medical Image Computing Platform

Jan Egger[1,2,3], Tina Kapur[1], Andriy Fedorov[1], Steve Pieper[1,4], James V. Miller[5], Harini Veeraraghavan[6], Bernd Freisleben[3], Alexandra J. Golby[1,7], Christopher Nimsky[2] & Ron Kikinis[1]

[1]Department of Radiology, Brigham and Women's Hospital, Harvard Medical School, Boston, MA, USA, [2]Department of Neurosurgery, University Hospital of Marburg, Marburg, Germany, [3]Department of Mathematics and Computer Science, The Philipps-University of Marburg, Marburg, Germany, [4]Isomics, Inc., Cambridge, MA, USA, [5]Interventional and Therapy Lab, GE Research, Niskayuna, NY, USA, [6]Biomedical Image Analysis Lab, GE Research, Niskayuna, NY, USA, [7]Department of Neurosurgery, Brigham and Women's Hospital, Harvard Medical School, Boston, MA, USA.

Volumetric change in glioblastoma multiforme (GBM) over time is a critical factor in treatment decisions. Typically, the tumor volume is computed on a slice-by-slice basis using MRI scans obtained at regular intervals. *(3D)Slicer* – a free platform for biomedical research – provides an alternative to this manual slice-by-slice segmentation process, which is significantly faster and requires less user interaction. In this study, 4 physicians segmented GBMs in 10 patients, once using the competitive region-growing based *GrowCut* segmentation module of *Slicer*, and once purely by drawing boundaries completely manually on a slice-by-slice basis. Furthermore, we provide a variability analysis for three physicians for 12 GBMs. The time required for *GrowCut* segmentation was on an average 61% of the time required for a pure manual segmentation. A comparison of *Slicer*-based segmentation with manual slice-by-slice segmentation resulted in a *Dice Similarity Coefficient* of 88.43 ± 5.23% and a *Hausdorff Distance* of 2.32 ± 5.23 mm.

Gliomas are the most common primary brain tumors, arising from the glial cells that support the cerebral nerve cells. The *World Health Organization (WHO)* grading system for gliomas defines grades I–IV, where grade I tumors are the least aggressive and IV are the most aggressive[1]. Of these, 70% are considered malignant gliomas (anaplastic astrocytoma *WHO* grade III and glioblastoma multiforme *WHO* grade IV). The glioblastoma multiforme, named for its histopathological appearance, is the most frequent malignant primary brain tumor and is one of the most highly malignant human neoplasms. The approach to the treatment of glioblastomas typically includes maximum safe resection, percutaneous radiation and chemotherapy. Despite new radiation strategies and the development of oral alcylating substances (e.g. Temozolomide), the life expectancy for GBM patients is still only about fifteen months[2]. Although in previous years the role of surgery was controversial, recent literature favors a maximum safe surgical resection as a positive predictor for extended patient survival[3]. Microsurgical resection can now be optimized with the technical development of neuronavigation based on data from diffusion tensor imaging (DTI), functional magnetic resonance imaging (fMRI), magnetoencephalography (MEG), magnetic resonance spectroscopy (MRS), or positron-emission-computed-tomography (PET). An early postoperative magnetic resonance imaging (MRI) with a contrast agent can be used to determine how much of the tumor mass has been removed and frequent MRI scans can help to monitor any new tumor growth.

For automatic glioma segmentation in general (*World Health Organization* grade I–IV), several algorithms have already been proposed that rely on magnetic resonance imaging. Szwarc et al.[4] have presented a segmentation approach that uses fuzzy clustering techniques. In their evaluation, the authors used six magnetic resonance (MR) studies of three subjects and the reported *Dice Similarity Coefficient (DSC)*[5,6] ranged from 67.21% to 75.63%. Angelini et al.[7] have presented an extensive overview of some deterministic and statistical approaches. Gibbs et al.[8] have introduced a combination of region growing and morphological edge detection for segmenting enhancing tumors in T1-weighted MRI data. The authors evaluated their method with one phantom data set and ten clinical data sets. An interactive method for segmentation of full-enhancing, ring-enhancing and non-enhancing tumors has been proposed by Letteboer et al.[9] and was evaluated on twenty clinical cases. Depending on intensity-based pixel probabilities for tumor tissue, Droske et al.[10] have presented a deformable model method, using a level set formulation, to divide the MRI data into regions of similar image properties for tumor segmentation. Clark et al.[11] have introduced a knowledge-based automated segmentation on multispectral data in order to partition

Table 1 | This table presents a comparison of a) the time it took for physicians to segment GBMs manually vs. using *3D Slicer*, b) the agreement between the two segmentations. The *MT* column shows the time (in minutes) it took a physician to segment each of ten GBMs on slice-by-slice basis. The *SlicerT* column shows the time (in minutes) it took a physician to segment it using *3D Slicer*. The *Slices* column shows the number of slices that the tumor spans in each case, as a rough approximation of the complexity of the segmentation task. Note that 9 out of 10 cases, *Slicer* < *MT*, and on an average, the time it took to segment with *3D Slicer* was 61% of the time it took to segment manually on a slice-by-slice basis. The columns *DSC* and *HD* show the agreement between the two segmentations using a *Dice Similarity Coefficient* and *Hausdorff Distance*, respectively

| Case No. | MT (min) | SlicerT (min) | Slices | SlicerT/MT | DSC | HD (mm) | Manual Volume (mm³) | Slicer Vol (mm³) | Slicer/Manual Vol |
|---|---|---|---|---|---|---|---|---|---|
| 1 | 9 | 4 | 36 | 0.44 | 0.85 | 2.80 | 33522 | 44694 | 1.33 |
| 2 | 19 | 7.5 | 51 | 0.39 | 0.91 | 3.68 | 28373 | 32383 | 1.14 |
| 3 | 6 | 4.5 | 42 | 0.75 | 0.92 | 1.71 | 42056 | 47752 | 1.14 |
| 4 | 16 | 6.5 | 60 | 0.41 | 0.91 | 3.00 | 69448 | 78776 | 1.13 |
| 5 | 3 | 2.5 | 10 | 0.83 | 0.81 | 2.00 | 1480 | 2016 | 1.36 |
| 6 | 14 | 6.25 | 43 | 0.45 | 0.94 | 2.00 | 39097 | 38905 | 1.00 |
| 7 | 13 | 8.5 | 36 | 0.65 | 0.87 | 2.23 | 22468 | 25331 | 1.13 |
| 8 | 7 | 9.25 | 42 | 1.32 | 0.92 | 2.12 | 27368 | 30648 | 1.12 |
| 9 | 5 | 3 | 11 | 0.60 | 0.79 | 2.39 | 2703 | 3908 | 1.45 |
| 10 | 11 | 2.5 | 16 | 0.23 | 0.92 | 0.31 | 10318 | 11720 | 1.14 |
| Averages | 10.30 | 5.45 | 34.70 | 0.61 | 0.88 | 2.32 | 27683 | 31613 | 1.19 |

{ Time } { Agreement }

glioblastomas. Direct comparison with a hand labeled segmentation 89 of 120 slices had a percent matching rate of 90% or higher. Segmentation based on outlier detection in T2-weighted MR data has been proposed by Prastawa et al.[12]. For each case, the time required for the automatic segmentation method was about ninety minutes. Sieg et al.[13] have introduced an approach to segment contrast-enhanced, intracranial tumors and anatomical structures of registered, multispectral MR data. The approach has been tested on twenty-two data sets, but no computation times were provided. Egger et al.[14,15] present a graph-based approach. After the graph has been constructed, the minimal cost closed set on the graph is computed via a polynomial time s-t cut[16]. The presented method has been evaluated with fifty glioblastoma multiforme yielding an average *Dice Similarity Coefficient* of 80.37 ± 8.93%.

Since fully automated segmentation often fails to match human judgments of tumor boundaries, a number interactive segmentation algorithms have been proposed. Vezhnevets and Konouchine[17] give an overview of methods for generic image editing and methods for editing medical images. An interactive segmentation technique called *Magic Wand*[17] is a common selection tool in image editing software applications. The tool gathers color statistics from the user specified image point (or region) and segments (connected) image regions with pixels whose color properties fall within some given tolerance of the gathered statistics. Reese[18] has presented a region-based interactive segmentation technique called *Intelligent Paint*, based on hierarchical image segmentation by tobogganing, with a connect-and-collect strategy to define an object's region. Mortensen and Barrett[19] have introduced a boundary-based method to compute a minimum-cost path between user-specified boundary points. The intelligent scissors method[20] treats each pixel as a graph node and uses shortest-path graph algorithms for boundary calculation and a faster variant of region-based intelligent scissors uses tobogganing for image over-segmentation and then treats homogenous regions as graph nodes. *GraphCut* is a combinatorial optimization technique applied to the task of image segmentation by Boykov and Jolly[21]. An extension of the *GraphCut* named *GrabCut* developed by Rother et al.[22], is an iterative segmentation scheme that uses a graph-cut for intermediate steps. A marker-based watershed transformation algorithm for medical image segmentation, developed by Moga and Gabbouj[23], uses user-specified markers for segmenting gray level images. The *Random Walker* algorithm of Grady and Funka-Lea[24] is a probabilistic approach using a small number of user-labeled pixels.

Heimann et al.[25] have presented an interactive region growing method that is a descendant of one of the classic image segmentation techniques. A manual refinement system for graph-based approaches has recently been presented by Egger et al.[26,27]. The approach takes advantage of the basic design of graph-based image segmentation algorithms and restricts a graph-cut by using additional user-defined seed points to set up fixed nodes in the graph. Another resent publication by Zukić et al.[28] presents semi-automatic GBM segmentation with a *balloon inflation* approach[29]. The balloon inflation method has been evaluated with twenty-seven magnetic resonance imaging data sets with a reported average *DSC* of 80.46%. The *GrowCut* method, developed by Vezhnevets and Konouchine[17], is a cellular automaton-based algorithm for interactive multilabel segmentation of *N*-dimensional images. The *GrowCut* algorithm is freely available as a module[30] for the medical image computing platform *3D Slicer*[31] and has been used in a recent study to segment Pituitary Adenomas[32].

In this paper, we present a detailed study of the volumetric analysis of glioblastoma multiforme using the *GrowCut* tool *3D Slicer*. Our objective is to evaluate the utility of *3D Slicer* in simplifying the time-consuming manual slice-by-slice segmentation while achieving a comparable accuracy. Thus, 4 physicians segmented GBMs in 10 patients, once using the competitive region-growing based *GrowCut* adopts uniform, rigorous response criteria similar to those in general oncology where response is defined as a ≥50% reduction in tumor size and the usual measure of "size" is the largest cross-sectional area (the largest cross-sectional diameter multiplied by the largest diameter perpendicular to it). Accurate and repeatable methods to calculate tumor volume are therefore an important aspect of clinical care.

Table 2 | Manual intra- and inter-physician segmentation results (min, max, mean μ and standard deviation σ) for three neurosurgeons – X, Y and Z – for twelve glioblastoma multiforme (GBM) data sets. The first column represents the intra-physician segmentation result: within a time distance of two weeks Physician *X* segmented the twelve GBMs slice-by-slice twice. The second and third columns present the inter-physician segmentation results, whereby the manual slice-by-slice segmentations form Physician *Y* and Physician *Z* have been compared with the first manual segmentation of Physician *X*

| | DSC for intra- and inter-physician segmentations | | |
|---|---|---|---|
| | Physician X | Physician Y | Physician Z |
| Min | 84.01% | 78.68% | 76.03% |
| Max | 96.30% | 94.86% | 94.83% |
| μ ± σ | 90.29 ± 4.48% | 88 ± 6.08% | 86.63 ± 6.87% |

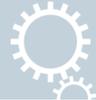
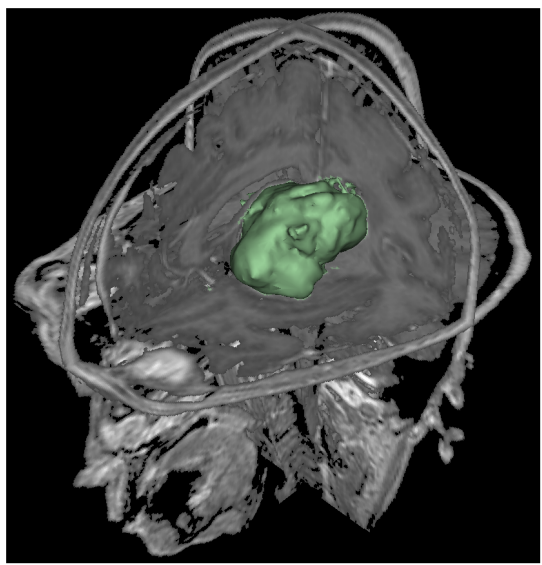

**Figure 1** | This image presents the segmentation results of *GrowCut* (green) for the tumor and background initialization of Figure 3. After the initialization of the *GrowCut* algorithm under *Slicer* it took about ten seconds to get the segmentation result on an *Intel Core i7-990 CPU, 12 × 3.47 GHz, 12 GB RAM, Windows 7 Home Premium x64 Version, Service Pack 1*.

The rest of this article is organized as follows: *Section 2* presents the results of our experiments. *Section 3* discusses the performance of the proposed approach, concludes the contribution and outlines areas for future work. Finally, *Section 4* presents the material and the methods.

## Results

The goal of this study was to evaluate the utility of *3D Slicer* for segmentation of GBMs compared to manual slice-by-slice segmentation. We used two metrics for this evaluation: a) the time it took for physicians to segment GBMs manually vs. using *3D Slicer*, b) the agreement between the two segmentations. In using these metrics to evaluate our results, our assumption is that if *3D Slicer* can be used to produce GBM segmentations that are statistically equivalent to what the physicians achieve manually, and in substantially less time, then the tool is useful for volumetric follow-ups of GBM patients. Overall, four physicians participated in our study: three physicians provided the manual slice-by-slice segmentations and one physician has been trained in a Slicer-based segmentation as described in the methods section. The results of our study are detailed in Table 1, the primary conclusion of which is that *3D Slicer* based GBM segmentation can be performed in about 60% of the time, and with acceptable agreement (DSC: 88.43 ± 5.23%, HD: 2.32 ± 5.23 mm) to manual segmentation by a qualified physician. In Table 1, The *MT* column shows the time (in minutes) it took a physician to segment each of ten GBMs on slice-by-slice basis. The *SlicerT* column shows the time (in minutes) it took a physician to segment it using *3D Slicer*. The *Slices* column shows the number of slices that the tumor spans in each case, as a rough approximation of the complexity of the segmentation task. Note that 9 out of 10 cases, *Slicer* < *MT*, and on an average, the time it took to segment with *3D Slicer* was **61%** of the time it took to segment manually on a slice-by-slice basis. The columns *DSC* and *HD* show the agreement between the two segmentations using a *Dice Similarity Coefficient* and *Hausdorff Distance*, respectively.

segmentation module of *3D Slicer*, and once by drawing boundaries manually on a slice-by-slice basis. The time required for *GrowCut* vs. manual segmentation were recorded. A comparison was performed of *3D Slicer* based segmentation with manual slice-by-slice segmentation using the *Dice Similarity Coefficient (DSC)* and the *Hausdorff Distance (HD)*[33–35].

Methods that use all slices to calculate the tumor boundaries have more information available to make accurate predictions of tumor volume. Simpler methods such as geometric models provide only a rough estimate of the tumor volume and may not be indicated for accurate determination of tumor burden. Geometric approximations use one or several user-defined diameters to estimate the tumor volume[36–38]. The Macdonald criteria[39] for measuring brain tumors

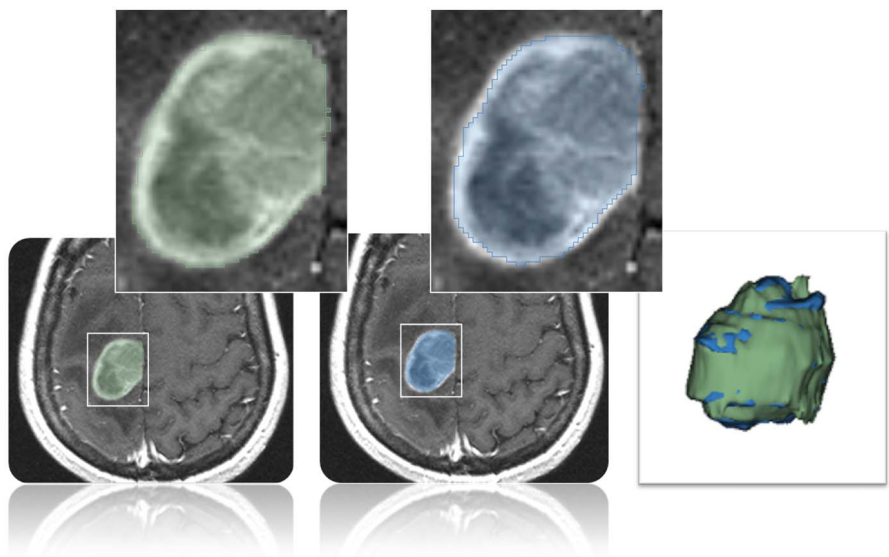

**Figure 2** | Comparison of glioblastoma multiforme (GBM) segmentation results on an axial slice: semi-automatic segmentation under *Slicer* (green, left image) and pure manual segmentation (blue, middle image). Moreover, a fused visualization of the 3D masks of the manual and the *Slicer* segmentation is presented (rightmost image).

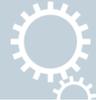

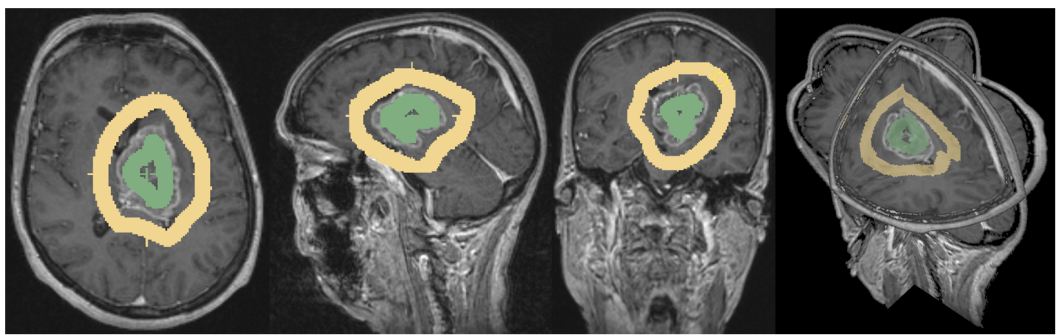

**Figure 3** | These images present a typical user initialization for glioblastoma multiforme (GBM) segmentation under *Slicer* with *GrowCut*: axial (left image) sagittal (second image from the left), and coronal (third image from the left). Besides, a 3D visualization of all three slices is presented (rightmost image). Note: the tumor has been initialized in green and the background has been initialized in yellow.

To provide readers with a point of comparison on how *DSC* and *HD* computations vary between expert raters, we include in Table 2 some statistics that we published in another article where we analyzed the results of 12 manual slice-by-slice GBM segmentations by 3 neurosurgeons[40,41].

In addition to the quantitative results, we present sample GBM segmentation results in Figures 1 and 2 for visual inspection. Figure 1 shows the results of the *3D Slicer GrowCut* function (for the tumor and background initialization shown in Figure 3). The rendered 3D tumor segmentation is superimposed (green) on three orthogonal cross-sections of the data. Figure 2 presents the direct comparison of *3D Slicer* vs. manual segmentation on an axial slice: the semi-automatic segmentation under *3D Slicer* (green) is shown on the left side of the figure and the pure manual segmentation (blue) is shown in the middle of the figure. A fused visualization of the 3D masks of the manual and the *Slicer* segmentations are displayed on the right side of the figure.

### Discussion

We observed that the automatic segmentation results produced by *3D Slicer* (*GrowCut*) typically required some additional editing on some slices to achieve the desired boundary and the time required for this manual correction is included in our measurements. Manual segmentation by neurosurgeons took three to nineteen minutes (mean: ten minutes), in contrast to the semi-automatic segmentation with the *GrowCut* implementation under *3D Slicer* that took about 60% of that time (mean: five minutes) including the time needed for editing the *GrowCut* results.

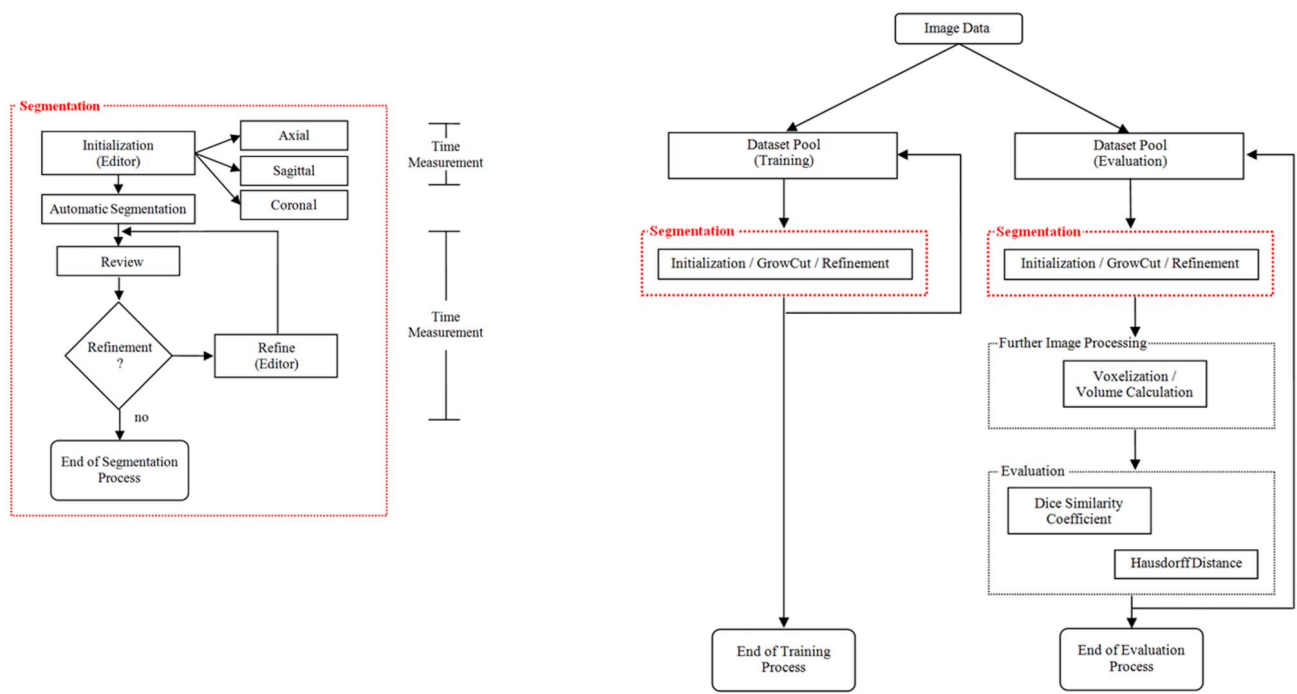

**Figure 4** | Detailed workflow of the segmentation process that is used in the training and the evaluation phase (left). The segmentation process starts with the initialization of the *GrowCut* algorithm by the user on an axial, sagittal and coronal slice. Then, the automatic segmentation is started and afterwards reviewed by the user. This results into the refinement phase where the *Editor* tools under *Slicer* are used to correct the automatic segmentation result – mostly by navigating along the axial slices. During the evaluation phase the time for the initialization and the refinement has been measured. The overall workflow of the proposed study is presented on the right side; it starts with the image data and ends with the training or the evaluation process. Therefore, the data is divided into two pools of data sets: the training data set and the evaluation data set. The segmentation process is for both stages the same. However, for the evaluation phase further image processing (voxelization and volume calculation) is required to calculate the *Dice Similarity Coefficient (DSC)* and the *Hausdorff Distance (HD)* for a quantitative evaluation.

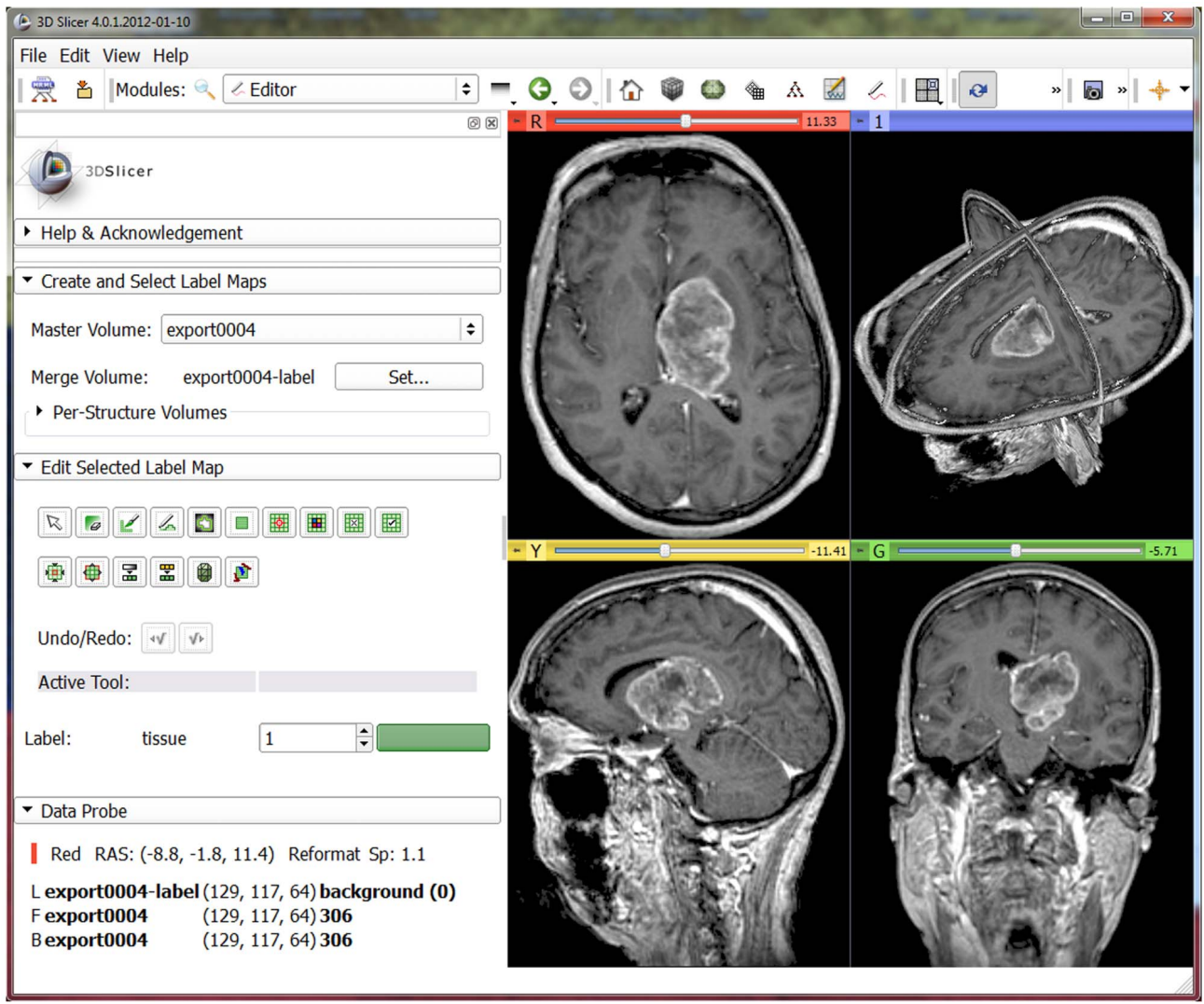

Figure 5 | *Slicer* interface with the *Editor* on the left side and a loaded glioblastoma multiforme (GBM) data set on the right side: axial slice (upper left window), sagittal slice (lower left window), coronal slice (lower right window) and the three slices shown in a 3D visualization (upper right window).

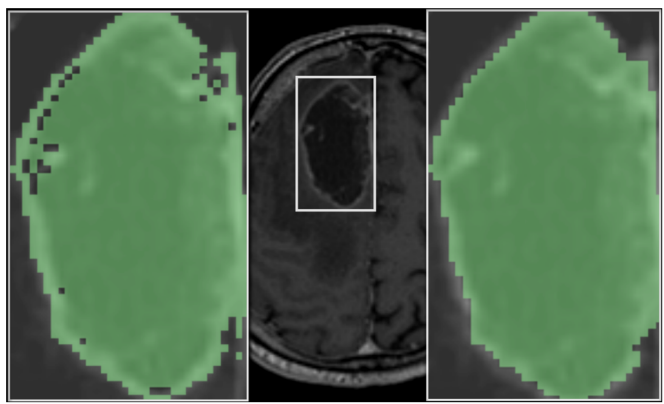

Figure 6 | **In these images the usage for the *Dilate* and *Erode* options under *Slicer* are presented.** The background shows an axial slice with a glioblastoma multiforme (white rectangle). The left white rectangle presents the zoomed segmentation result of *GrowCut* (green). As shown, the segmentation result is not very smooth at the tumor border. To get a smoother result the *Dilate* and *Erode* options under *Slicer* can be used. For this example *Dilate*, *Erode* and an additional *Erode* have been performed. The result of this operations is shown in the right white rectangle (green).

To quantify the quality of the *GrowCut* algorithm, we performed intra- and inter-physician segmentations[40,41]. The results also provided an upper segmentation threshold and therefore a quality measure for our algorithm. For the intra-physician segmentation, a neurosurgeon segmented twelve glioblastoma multiforme. After two weeks, the same neurosurgeon segmented these twelve cases again. The detailed results are presented in Table 2 and provide a mean value $\mu$ and a standard deviation $\sigma$ of 90.29 ± 4.48% with a minimal *Dice Similarity Coefficient* of 84.01% and a maximal *Dice Similarity Coefficient* of 96.30% (see the first column). Finally, Table 2 also shows inter-physician segmentation results for the twelve glioblastoma multiforme (see the second and third columns). Therefore, the segmentation of the neurosurgeons Y and Z have been compared with the segmentations of neurosurgeon X. It is evident that there is an upper threshold with a *Dice Similarity Coefficient* of around ninety percent for the manual intra- and inter-physician segmentations (average *DSC* when compared with an automatic segmentation: 79.96 ± 8.06% (neurosurgeon X), 77.79 ± 8.49% (neurosurgeon Y) and 76.83 ± 13.67% (neurosurgeon Z)). The DSC of 90% can be thought of as a metric for estimating how well an automatic segmentation result is performing relative to the range of performance of experts, and perhaps also can serve as an indicator for how much manual post-editing will be required after the automatic segmentation is performed.

In this paper, the evaluation of glioblastoma multiforme segmentation with the free and open source medical image analysis software *3D Slicer* has been presented. *Slicer* provides a semi-automatic, 3D segmentation algorithm, *GrowCut*, that is a viable alternative to the time-consuming process of volume determination during monitoring of a patient, for which slice-by-slice contouring has been the best demonstrated practice. Editing tools available in *3D Slicer* are used for manual editing of the results upon completion of the automatic *GrowCut* segmentation. The volume of the 3D tumor is then computed and stored as an aide for the surgeon in decision making for comparison with follow-up scans. This segmentation has been evaluated on 10 GBM data sets against manual expert segmentations using the *Dice Similarity Coefficient (DSC)* and the *Hausdorff Distance (HD)*. Additionally, intra-physician segmentations have been performed to provide a quality measure of the presented evaluation. In summary, the achieved research highlights of the presented work are:

- Manual slice-by-slice segmentations of glioblastoma multiforme (GBM) have been performed by clinical experts to obtain ground truth of tumor boundaries and estimates of rater variability.
- Physicians have been trained in segmenting glioblastoma multiforme with *GrowCut* and the *Editor* module of *3D Slicer*.
- Trained physicians used *Slicer* to segment a glioblastoma multiforme evaluation set.
- Semi-automatic segmentation times have been measured for *GrowCut* based segmentation in *3D Slicer*.
- *Dice Similarity Coefficient (DSC)* and *Hausdorff Distance (HD)* have been calculated to evaluate the quality of the segmentations.

There are several areas for future work. In particular, some steps of the segmentation workflow under *Slicer* can be automated. Instead of initializing the foreground on three single 2D slices, a single 3D initialization could be used by means of generating a sphere around the position of the user-defined seed point. Additionally, the algorithm can be enhanced with statistical information about the shape[42] and the texture of the desired object[43] to improve the automatic segmentation. Furthermore, we plan to evaluate the method on magnetic resonance imaging (MRI) data sets with *World Health Organization* grade I, II and III gliomas. As compared to high-grade gliomas, low-grade tumor MR images lack gadolinium enhancement. Thus, for these tumors, outlines cannot be expressed by contrast-enhancing T1-weighted images, but by surrounding edema in T2-weighted images. In addition, we want to study how *Slicer* can be used to enhance the segmentation process of vertebral bodies. Besides, we want to apply the scheme to segment other organs and pathologies. Moreover, we are considering improving the algorithm by performing the whole segmentation iteratively; that is, after the segmentation has been performed, the result of the segmentation can be used as a new initialization for a new segmentation run with the process repeated under user control. We anticipate that the iterative approach will result in more robustness with respect to initialization.

## Methods

**Data.** Ten diagnostic T1-weighted MRI scans with gadolinium enhancement were used for segmentation. These were acquired on a 1.5 Tesla MRI scanner (Siemens MAGNETOM Sonata, Siemens Medical Solutions, Erlangen, Germany) using a standard head coil. Scan parameters were: TR/TE 2020/4.38 msec, isotropic matrix, 1 mm; FOV, 250 × 250 mm; 160 sections.

**Software.** For the semi-automatic segmentation work in this study we used *3D Slicer* 4.0, which is freely downloadable from the website http://www.slicer.org.

Manual segmentation of each data set was performed on a slice-by-slice basis by neurosurgeons at the *University Hospital of Marburg in Germany* (Chairman: Prof. Dr. Ch. Nimsky) with several years of experience in the resection of gliomas (note: if the tumor border was very similar between consecutive slices, the software allowed the user to skip manual segmentation in each slice, and instead interpolated the boundaries in these areas). The software used for this manual contouring provided simple contouring capabilities, and was created by us using the medical prototyping platform *MeVisLab* (see http://www.mevislab.de/). The hardware platform used was an *Intel Core i5-750 CPU, 4 × 2.66 GHz, 8 GB RAM, Windows XP Professional ×64 Version, Version 2003, Service Pack 2*.

**GrowCut segmentation in 3D Slicer.** The *GrowCut* is an interactive segmentation algorithm based on the idea of cellular automaton. The algorithm achieves reliable and reasonably fast segmentation of moderately difficult objects in 2D and 3D using an iterative labeling procedure resembling competitive region growing. A user's interactions results in a set of seed pixels which in turn try to assign their labels to their pixel neighborhood. A pixel is assigned the label of its neighbor when the similarity measure of the two pixels weighted by the neighboring pixel's weight or "strength" exceeds its current weight. Label assignment also results in an update of the pixel's weight. The labeling procedure continues iteratively until a stable configuration is reached when modification of the pixel labels is no longer possible. The algorithm is simple to use requiring no additional inputs from the user besides the painted strokes on the apparent foreground and background. Furthermore, the user can modify the segmentation by adding additional labels in the image, thereby influencing the segmentation result.

Our implementation of the algorithm in *3D Slicer* consists of a GUI front-end to enable interactions of the user with the image and an algorithm back-end where the segmentation is computed. We employ a minimal interface, where the user interacts by painting on the image. The algorithm requires labeling with at least two different colors (for a foreground and a background label class). The naïve implementation of the algorithm would require every pixel to be visited in each iteration. Furthermore, a pixel will need to visit every one of its neighbors to update the pixel strengths and labels. Such an implementation would be computationally expensive especially for large 3D images. We implemented the following techniques for speeding up the segmentation. First, as the user may be interested only in segmenting out a small area in the image, the algorithm computes the segmentation only within a small *region of interest (ROI)*. The ROI is computed as a convex hull of all user labeled pixels with an additional margin of approximately 5% for our study. Second, the iterations involving the image are executed in multiple threads, such that several small regions of the image are updated simultaneously (note: the implementation is multithreaded and automatically makes use of all the cores of the computer). Finally, the similarity distance between the pixels are pre-computed once and reused. Also the algorithm keeps track of *saturated* pixels (those whose weights and therefore labels can no longer be updated) and avoids the expensive neighborhood computation on those pixels. Keeping track of such pixels also helps to determine when to terminate the algorithm.

**GBM segmentation using 3D Slicer.** After trials of the various segmentation facilities available in *Slicer*, we determined that the use of *GrowCut* followed by morphological operations such as *erosion*, *dilation*, and *island removal* provides the most efficient segmentation method for GBMs from gadolinium enhanced T1 images. As shown in Figure 4, we used the following workflow to perform GBM segmentation: 1) load the data set into *Slicer* 2) initialization of an area inside the tumor, and a stroke drawn outside the tumor with a brush size of about 1 cm 3) automatic competing region-growing using *GrowCut*, and 4) usage of Editing tools like *dilation*, *erosion*, and *island removal* or pure manual refinement after visual inspection of results (note: the users are responsible for qualitatively deciding how much dilation, erosion and island removal are required for the segmentation). Figure 5 shows the *Slicer Editor* module user interface on the left side and a loaded GBM data set on the right side. Figure 3 presents a typical user initialization for *GrowCut* on the axial, sagittal and coronal cross-sections. Figure 6 shows the results of subsequent *erosion* followed by a *dilation*, and Figure 1 shows the results of the *GrowCut* method.

The hardware platform used was an Apple MacBook Pro (4 Intel Core i7, 2.3 GHz, 8 GB RAM, AMD Radeon HD 6750 M, Mac OS × 10.6 Snow Leopard).

**Measurement of segmentation time.** We measured the time taken by the same physician to segment manually vs. the *3D Slicer* method. Within the *3D Slicer* segmentation, we separately measured the time taken by each of the three steps (initialization, *GrowCut*, refinement using morphological operations) of the *3D Slicer* method (see left chart of Figure 4).

**Metrics for comparison between 3D Slicer and manual segmentation.** The resulting segmentations from both methods were saved as binary volumes, and the agreement between the two was compared using the *Dice Similarity Coefficient* and the *Hausdorff Distance*.

The *Dice Similarity Coefficient (DSC)* of agreement between two binary volumes is calculated as follows:

$$DSC = \frac{2 \cdot V(A \cap R)}{V(A) + V(R)} \quad (1)$$

The *DSC* measures the relative volume overlap between *A* and *R*, where *A* and *R* are the binary masks from the automatic (*A*) and the reference (*R*) segmentation. $V(\cdot)$ is
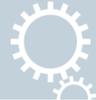

the volume (in mm³) of voxels inside the binary mask, by means of counting the number of voxels, then multiplying with the voxel size.

The *Hausdorff Distance (HD)* between two binary volumes is defined in terms of the *Euclidean* distance between the boundary voxels of the masks. Given the sets *A* (of the automatic segmentation) and *R* (of the reference segmentation) that consist of the points that correspond to the centers of segmentation mask boundary voxels in the two images, the directed *HD h(A,R)* is defined as the minimum *Euclidean* distance from any of the points in the first set to the second set, and the *HD* between the two sets *H(A,R)* is the maximum of these distances:

$$h(A,R) = \max_{a \in A}(d(a,R)), \text{ where } d(a,R) = \min_{r \in R}\|a-r\| \quad (2)$$

$$H(A,R) = \max(h(A,R), h(R,A))$$

1. Kleihues, P. *et al.* The WHO classification of tumors of the nervous system. *Journal of Neuropathology & Experimental Neurology* **61**(3), 215–229 (2002).
2. Kortmann, R. D., Jeremic, B., Weller, M., Plasswilm, L. & Bamberg, M. Radiochemotherapy of malignant gliom in adults. Clinical experiences. *Strahlenther Onkol.* **179**(4), 219–232 (2003).
3. Lacroix, M. *et al.* A multivariate analysis of 416 patients with glioblastoma multiforme: Prognosis, extent of resection and survival. *Journal of Neurosurgery* **95**, 190–198 (2001).
4. Szwarc, P., Kawa, J., Bobek-Billewicz, B. & Pietka, E. Segmentation of Brain Tumours in MR Images Using Fuzzy Clustering Techniques. Proceedings of Computer Assisted Radiology and Surgery (CARS), Geneva, Switzerland (2010).
5. Zou, K. H. *et al.* Statistical Validation of Image Segmentation Quality Based on a Spatial Overlap Index: Scientific Reports. *Academic Radiology* **11**(2), 178–189 (2004).
6. Sampat, M. P. *et al.* Measuring intra- and inter-observer agreement in identifying and localizing structures in medical images. *IEEE Inter Conf Image Processing*, 4 pages (2006).
7. Angelini, E. D. *et al.* Glioma Dynamics and Computational Models: A Review of Segmentation, Registration, and In Silico Growth Algorithms and their Clinical Applications. *Current Med. Imaging Rev.* **3**, 262–76 (2007).
8. Gibbs, P. *et al.* Tumour volume determination from MR images by morphological segmentation. *Physics in Med. & Biology* **41**(11), 2437–46 (1996).
9. Letteboer, M. M. J. *et al.* Segmentation of tumors in magnetic resonance brain images using an interactive multiscale watershed algorithm. *Academic Radiology* **11**, 1125–1138 (2004).
10. Droske, M. *et al.* An adaptive level set method for interactive segmentation of intracranial tumors. *Neurol Res* **27**(4), 363–70 (2005).
11. Clark, M. *et al.* Automatic tumor segmentation using knowledge-based techniques. *IEEE Transactions on Medical Imaging (TMI)* **17**(2), 187–201 (1998).
12. Prastawa, M. *et al.* A brain tumor segmentation framework based on outlier detection. *Medical Image Analysis* **8**, 275–283 (2004).
13. Sieg, C., Handels, H. & Pöppl, S. J. Automatic Segmentation of Contrast-Enhanced Brain Tumors in Multispectral MR-Images with Backpropagation-Networks (in German). Bildverarbeitung für die Medizin (BVM), Springer Press, 347–351 (2001).
14. Egger, J. *et al.* Nugget-Cut: A Segmentation Scheme for Spherically- and Elliptically-Shaped 3D Objects. 32nd Annual Symposium of the German Association for Pattern Recognition (DAGM). *LNCS* **6376**, 383–392, Springer Press, Darmstadt, Germany (2010).
15. Egger, J., Kappus, C., Freisleben, B., & Nimsky, C. A Medical Software System for Volumetric Analysis of Cerebral Pathologies in Magnetic Resonance Imaging (MRI) Data. *J Med Syst.* 2012 Aug **36**(4), 2097–10 (2012).
16. Boykov, Y. & Kolmogorov, V. An Experimental Comparison of Min-Cut/Max-Flow Algorithms for Energy Minimization in Vision. *IEEE Transactions on Pattern Analysis and Machine Intelligence* **26**(9), 1124–1137 (2004).
17. Vezhnevets, V. & Konouchine, V. GrowCut-Interactive multi-label N-D image segmentation. *Proc. Graphicon*, 150–156 (2005).
18. Reese, L. Intelligent Paint: Region-Based Interactive Image Segmentation. Master's thesis, Department of Computer Science, Brigham Young University, Provo, UT (1999).
19. Mortensen, E. N. & Barrett, W. A. Interactive segmentation with intelligent scissors. *Graphical Models and Image Processing* **60**(5), 349–384 (1998).
20. Mortensen, E. N. & Barrett, W. A. Toboggan-based intelligent scissors with a four-parameter edge model. *IEEE Conference on Computer Vision and Pattern Recognition (CVPR)*, pages 2452–2458 (1999).
21. Boykov, Y. & Jolly, M.-P. Interactive graph cuts for optimal boundary and region segmentation of objects in n-d images. *Proceedings of the International Conference on Computer Vision (ICCV)*, volume **1**, pages 105–112 (2001).
22. Rother, C., Kolmogorov, V. & Blake, A. Grabcut-interactive foreground extraction using iterated graph cuts. *Proceedings of ACM Siggraph* (2004).
23. Moga, A. & Gabbouj, M. A parallel marker based watershed transformation. *In IEEE International Conference on Image Processing (ICIP)*, **II**, 137–140 (1996).
24. Grady, L. & Funka-Lea, G. Multi-label image segmentation for medical applications based on graph-theoretic electrical potentials. *ECCV Workshops CVAMIA and MMBIA*, 230–245 (2004).
25. Heimann, T., Thorn, M., Kunert, T. & Meinzer, H.-P. New methods for leak detection and contour correction in seeded region growing segmentation. In 20th ISPRS Congress, Istanbul 2004. *International Archives of Photogrammetry and Remote Sensing,* vol. **XXXV**, 317–322 (2004).
26. Egger, J., Colen, R., Freisleben, B. & Nimsky, C. Manual Refinement System for Graph-Based Segmentation Results in the Medical Domain. *J Med Syst.* 2012 Oct **36**(5), 2829–39 (2011).
27. Egger, J. *et al.* A Flexible Semi-Automatic Approach for Glioblastoma multiforme Segmentation. Proceedings of International Biosignal Processing Conference, Charité, Berlin, Germany, 4 pages (2010).
28. Zukić, D. *et al.* Glioblastoma Multiforme Segmentation in MRI Data with a Balloon Inflation Approach. Proceedings of 6th RB Conference on Bio-Medical Engineering, State Technical University, Moscow, Russia, 4 pages (2010).
29. Cohen, L. D. On active contour models and balloons. *CVGIP: Image Understanding* **53**(2), 211–218 (1991).
30. Egger, J. *et al.* Square-Cut: A Segmentation Algorithm on the Basis of a Rectangle Shape. *PLoS One* **7**(2), e31064. Epub Feb 21 (2012).
31. Fedorov, A. *et al.* 3D Slicer as an Image Computing Platform for the Quantitative Imaging Network. *Magnetic Resonance Imaging 2012;* July PMID: 22770690 (2012).
32. Egger, J., Kapur, T., Nimsky, C. & Kikinis, R. Pituitary Adenoma Volumetry with 3D Slicer. *PLoS ONE* **7**(12), e51788 (2012).
33. Hausdorff, F. Grundzüge der Mengenlehre. Veit & Comp., Leipzig 1914 (rep. in Srishti D. Chatterji et al. (Hrsg.), Gesammelte Werke, Band II, Springer, Berlin, ISBN 3-540-42224-2 (2002).
34. Rockafellar, R. T. & Wets, R. J.-B. Variational Analysis. Springer, page 117, ISBN 3540627723, ISBN 978-3540627722 (2005).
35. Huttenlocher, D. P., Klanderman, G. A. & Rucklidge, W. J. Comparing images using the Hausdorff distance. *IEEE Transactions on Pattern Analysis and Machine Intelligence* **15**(9), 850–863 (1993).
36. Jimenez, C. *et al.* Follow-up of pituitary tumor volume in patients with acromegaly treated with pegvisomant in clinical trials. *European Journal of Endocrinology* **159**, 517–523 (2008).
37. Nobels, F. R. E. *et al.* Long-term treatment with the dopamine agonist quinagolide of patients with clinically non-functioning pituitary adenoma, in European. *Journal of Endocrinology* **143**, 615–621 (2000).
38. Korsisaari, N. *et al.* Blocking Vascular Endothelial Growth Factor-A Inhibits the Growth of Pituitary Adenomas and Lowers Serum Prolactin Level in a Mouse Model of Multiple Endocrine Neoplasia Type 1. *Clinical Cancer Research, January 1* **14**(1), 249–258 (2008).
39. Macdonald, D. R., Cascino, T. L., Schold, S. C. Jr. & Cairncross, J. G. Response criteria for phase II studies of supratentorial malignant glioma. *J Clin Oncol.* **8**(7), 1277–80 (1990).
40. Egger, J., Bauer, M. H. A., Kuhnt, D., Freisleben, B. & Nimsky, C. Min-Cut-Segmentation of WHO Grade IV Gliomas Evaluated against Manual Segmentation. XIX Congress of the European Society for Stereotactic and Functional Neurosurgery, Athens, Greece (2010).
41. Egger, J. *et al.* Evaluation of a Novel Approach for Automatic Volume Determination of Glioblastomas Based on Several Manual Expert Segmentations. Proceedings of 44. Jahrestagung der DGBMT, Rostock, Germany (2010).
42. Cootes, T. F. & Taylor, C. J. Active Shape Models-'Smart Snakes'. *Proceedings of the British Machine Vision Conference,* pages 266–275(1992).
43. Cootes, T. F. & Taylor, C. J. Statistical Models of Appearance for Computer Vision. Technical report, University of Manchester (2004).

### Acknowledgements
The authors would like to thank the physicians: Dr. med. Barbara Carl, Rivka Colen, M.D., Christoph Kappus and Dr. med. Daniela Kuhnt for performing the manual segmentations of the medical images and participating in the study. The authors would also like to thank the members of the *Slicer Community* for their contributions. Furthermore, the authors would like to thank *Fraunhofer MeVis* in Bremen, Germany, for their collaboration and especially Prof. Dr. Horst K. Hahn for his support.
This project was supported by the National Center for Research Resources (P41RR019703) and the National Institute of Biomedical Imaging and Bioengineering (P41EB015898, U54EB005149, R03EB013792) of the National Institutes of Health. Its contents are solely the responsibility of the authors and do not necessarily represent the official views of the NIH.

### Author contributions
Conceived and designed the experiments: J.E. & A.F. Performed the experiments: J.E. & A.F. Analyzed the data: J.E. & A.F. Contributed reagents/materials/analysis tools: J.E., T.K., A.F., S.P., J.V.M., H.V., B.F., A.J.G., C.N. & R.K. Wrote the paper: J.E., T.K. & S.P.

### Additional information

**How to cite this article:** Egger, J. *et al.* GBM Volumetry using the 3D Slicer Medical Image Computing Platform. *Sci. Rep.* **3**, 1364; DOI:10.1038/srep01364 (2013).